\documentclass{article}

\usepackage{microtype}
\usepackage{graphicx}
\usepackage{subcaption}
\usepackage{booktabs} % for professional tables

\usepackage{hyperref}

\usepackage[preprint]{icml2026}

\usepackage{amsmath}
\usepackage{amssymb}
\usepackage{mathtools}
\usepackage{amsthm}
\usepackage{booktabs}
\usepackage{multirow}
\usepackage{makecell}
\usepackage[table]{xcolor}

\usepackage{arydshln}%

\usepackage[capitalize,noabbrev]{cleveref}

\theoremstyle{plain}

\theoremstyle{definition}

\theoremstyle{remark}

\usepackage[textsize=tiny]{todonotes}

\icmltitlerunning{Where to Refine, When to Stop: Rethinking Redundancy via Latent Discrepancy
for Efficient Visual Autoregressive Generation}

\begin{document}

\twocolumn[
  \icmltitle{Where to Refine, When to Stop: Rethinking Redundancy via Latent Discrepancy for Efficient Visual Autoregressive Generation}
  % Where to Refine, When to Stop: Rethinking Redundancy via Texture and Semantic Complexity for Efficient Visual Autoregressive Generation
  % Where to Refine, When to Stop: Rethinking Redundancy via Texture and Semantic Complexity in Visual Autoregressive Modeling
  % Where to Refine, When to Stop: Accelerating Visual Autoregressive Generation via Latent Spectral Analysis
  % EfficientVAR: Accelerating Visual Autoregressive Generation via Latent High-Frequency Pruning and Semantic-Adaptive Guidance
  % SHARP-VAR: Sharpening Visual Autoregressive Efficiency via Explicit Signal Pruning



  \icmlsetsymbol{equal}{*}

  \begin{icmlauthorlist}
    \icmlauthor{Changwang Mei}{yyy,comp,equal}
    \icmlauthor{Peisong Wang}{comp,equal}
    \icmlauthor{Zekun Li}{comp,baai}
    \icmlauthor{Changsheng Li}{sch}
    \icmlauthor{Shuang Qiu}{cuhk}
    \icmlauthor{Qinghao Hu}{comp}
    \icmlauthor{Gang Li}{comp}
    \icmlauthor{Yifan Zhang}{comp}
    \icmlauthor{Zhihui Wei}{yyy}
    \icmlauthor{Jian Cheng}{comp}
  \end{icmlauthorlist}

  \icmlaffiliation{yyy}{Nanjing University of Science and Technology}
  \icmlaffiliation{comp}{$\mathrm{C}^2\mathrm{DL}$, Institute of Automation, Chinese Academy of Sciences}
  \icmlaffiliation{sch}{Beijing Institute of Technology}
  \icmlaffiliation{baai}{Beijing Academy of Artificial Intelligence}
  \icmlaffiliation{cuhk}{Department of Systems Engineering and Department of Computer Science, City University of Hong Kong}
  \icmlsetsymbol{equal}{*}

  \icmlcorrespondingauthor{Zhihui Wei}{gswei@njust.edu.cn}
  \icmlcorrespondingauthor{Jian Cheng}{jcheng@nlpr.ia.ac.cn}

  \icmlkeywords{Visual Autoregressive Models, Efficient Image Generation, Training-free Acceleration, Token Pruning, Deep Learning}

  \vskip 0.3in
]



\printAffiliationsAndNotice{\icmlEqualContribution}

\begin{abstract}
Visual Autoregressive (VAR) models deliver high-quality image generation but suffer from significant inference latency at high resolutions. Recent acceleration approaches most rely on heuristic measures with layer features to prune tokens. Such heuristics are sensitive to complex contextual semantics, leading to inaccurate identification of redundant computation and poor adaptability across prompts.
We rethink redundancy in VAR from the perspective of its impact on pixel-space generation and introduce \textbf{Latent Discrepancy}. This unified metric quantifies a token’s contribution by measuring the change in model states during generation. Our analysis shows that redundancy is more accurately identified when guided by image latent or pixel-space signals. We further observed that in classifier-free guidance (CFG), the convergence trend of the discrepancy between conditional and unconditional branches exhibits high dynamics with different prompts.
Based on these findings, we propose \textbf{LD-Pruning} (\textbf{L}atent \textbf{D}iscrepancy \textbf{Pruning}), a training-free framework that removes redundancy via latent discrepancy by integrating decoding-free region selection and adaptive unconditional-branch skipping. Extensive experiments show that LD-Pruning substantially reduces inference latency while maintaining \textbf{high generation quality}, achieving up to \textbf{2.35$\times$} speedup on \textbf{Infinity-8B}.

\end{abstract}

\begin{figure}[t]
  \begin{center}\centerline{\includegraphics[width=0.9\columnwidth]{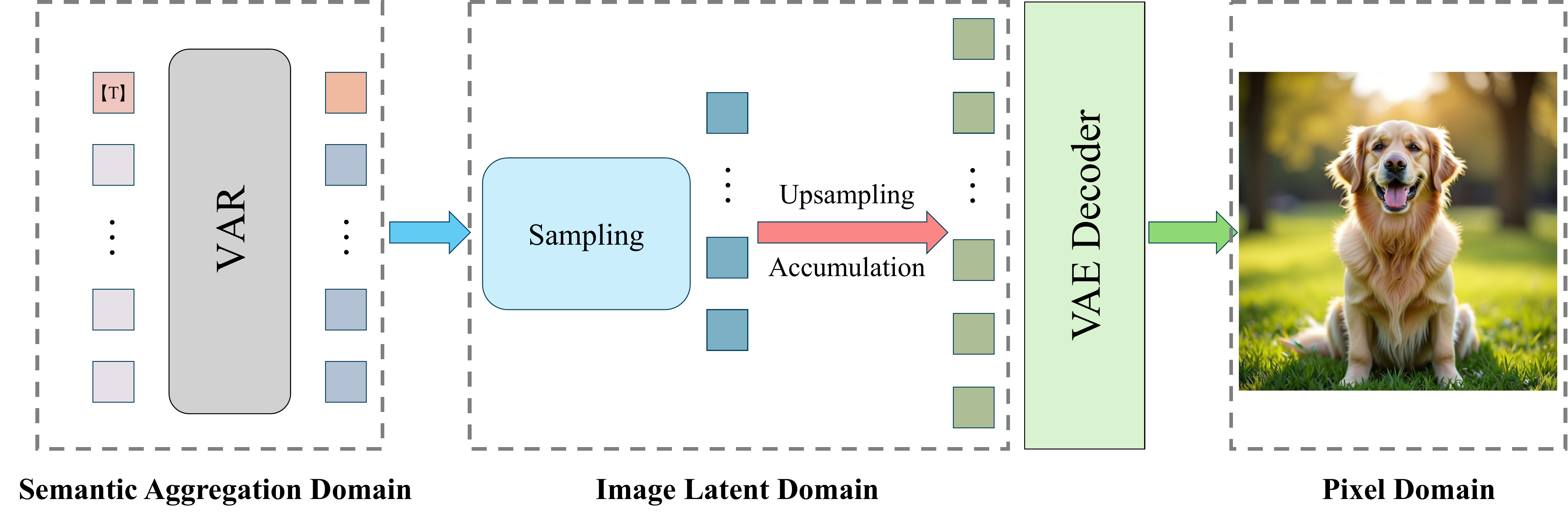}}
    \caption{Overview of the VAR inference pipeline. The Semantic Aggregation Domain integrates abstract text semantics and contextual information to generate features. These features are then converted into image-aligned latent representations within the Image Latent Domain. Finally, the Pixel Domain decodes these representations into the resulting image.
    }
    \label{domain}
  \end{center}
  \vskip -0.4in
\end{figure}

\begin{figure*}[t]
  % \vskip 0.2in
  \begin{center}
    \centerline{\includegraphics[width=0.94\textwidth]{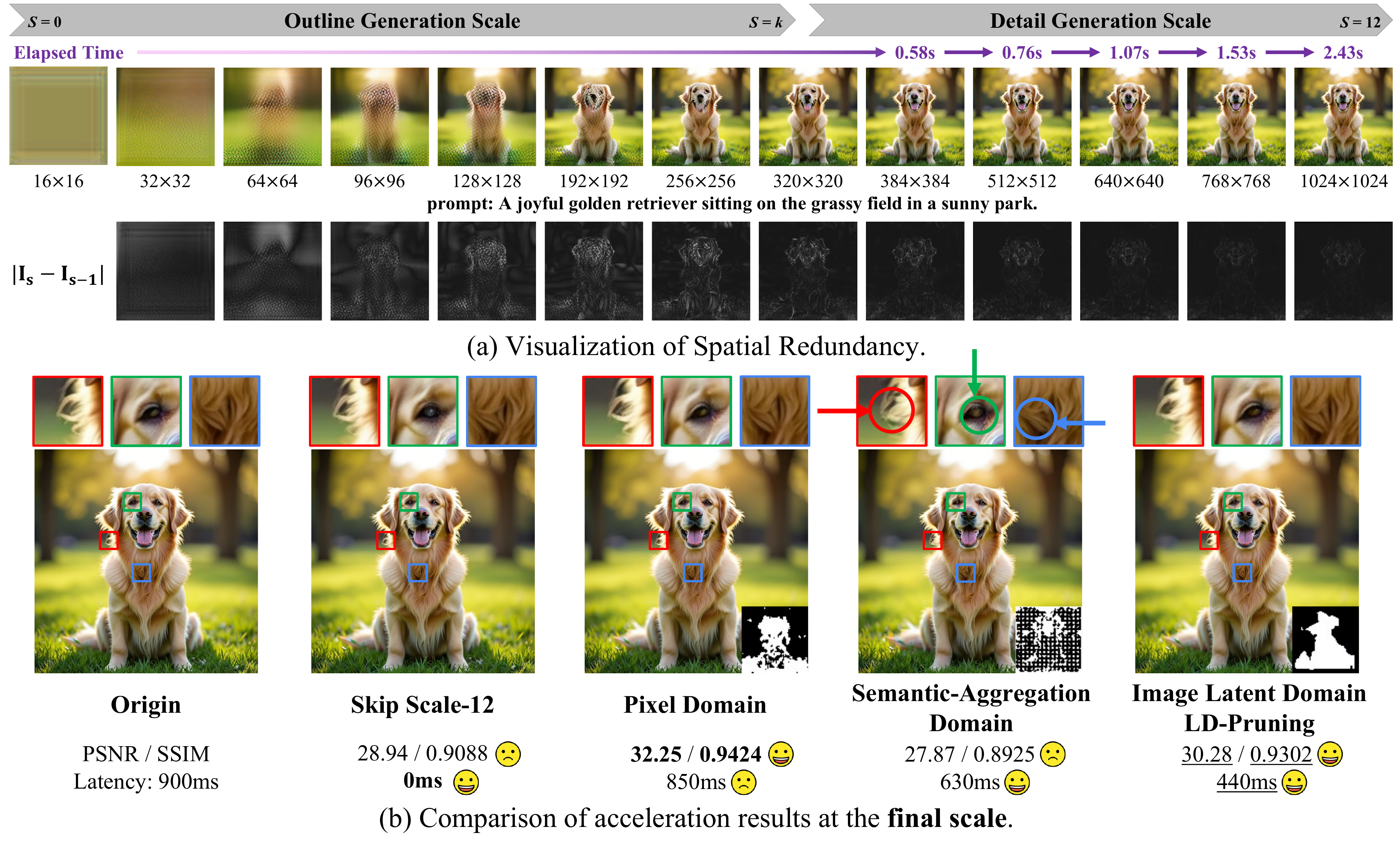}}
    \caption{
      Analysis of spatial redundancy in VAR and comparison of acceleration schemes. (a) Sparse pixel updates ($|I_k - I_{k-1}|$, where $I_k$ is the decoded image at scale $k$) in the detail-generation scales suggest later stages mainly refine high-frequency textures, revealing spatial redundancy; all latency results are measured on a single RTX 3090 GPU. (b) \textbf{LD-Pruning} uses decoding-free latent high-frequency energy to preserve fine details (e.g., fur) comparable to pixel-domain signals, while significantly accelerating inference.
    }
    \label{Spatial Redundancy}
  \end{center}
  \vskip -0.2in
\end{figure*}

\begin{figure*}[t]
  % \vskip 0.2in
  \begin{center}
    \centerline{\includegraphics[width=1.\textwidth]{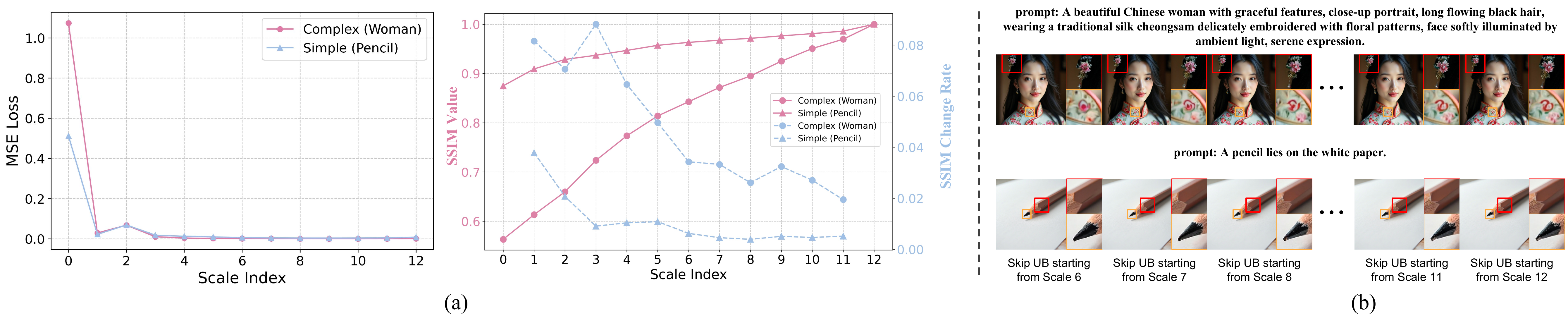}}
    \caption{
        Analysis of Guidance Redundancy. (a) Semantic-dependent convergence analysis. Left: conditional--unconditional discrepancy measured by MSE across scales. Right: generation quality under
        different unconditional-branch skipping scales, measured by SSIM and its change rate. (b) Generation results of skipping the unconditional branch (UB) starting from different scales.
    }
    \label{Guidance Redundancy}
  \end{center}
  \vskip -0.2in
\end{figure*}

\section{Introduction}

Autoregressive (AR) models \cite{lee2022autoregressive,liu2024lumina,li2024autoregressive} have unified visual understanding and generation \cite{shi2024lmfusion,sun2024autoregressive,qu2025tokenflow,wu2025janus} but suffer from prohibitive latency due to sequential token-by-token decoding. Visual Autoregressive (VAR) modeling addresses this by reframing generation as a coarse-to-fine next-scale prediction \cite{tian2024visual,tang2024hart,han2025infinity}, achieving diffusion-level zero-shot capabilities. However, despite this paradigm shift, VAR still faces substantial computational overhead as image resolution increases, creating a severe bottleneck for real-time applications.

To address this issue, recent acceleration methods achieve substantial speedups via two routes:
(i) pruning/approximating spatial tokens to reduce per-step computation, and
(ii) skipping the unconditional branch in classifier-free guidance (CFG) to cut guidance overhead.
However, both are still limited.

\noindent\textbf{(1) Token selection for accelerated inference.}
As shown in Figure~\ref{domain}, VAR inference consists of the \textbf{Semantic Aggregation Domain},
the \textbf{Image Latent Domain}, and the \textbf{Pixel Domain}.
Most prior work performs token selection primarily based on signals in the \textbf{Semantic Aggregation Domain}.
Concretely, FastVAR~\cite{guo2025fastvar} retains Top-$K$ tokens ranked by high-frequency responses and fills pruned positions with cached/interpolated tokens, while StageVAR~\cite{li2025stagevar} keeps representative tokens (selected in a low-rank subspace) and reconstructs the rest from cached/upsampled features.
\textbf{Limitation 1:} Due to interference from complex contextual semantics, they fail to identify redundant tokens accurately, leading to significant degradation in image textures and fine details (see Figure~\ref{Spatial Redundancy}(b), more details in section \ref{sec:rethinking spatial redundancy} ). 

\noindent\textbf{(2) Static unconditional-branch skipping in CFG.}
Another line accelerates sampling by reducing CFG cost through unconditional-branch skipping/replacement, but typically with static schedules.
For example, SkipVAR~\cite{li2025skipvar} makes a fixed decision at a pre-defined point (step 9) and applies unconditional-branch replacement
only to a pre-defined set of late steps (typically 10 to 12), reusing the conditional output to bypass the unconditional branch.
\textbf{Limitation 2:} Such static strategies lack a reliable online, prompt-adaptive criterion for when guidance becomes unnecessary,
which may terminate guidance too early for semantically complex prompts or incur unnecessary computation for simpler ones
(see Figure~\ref{Guidance Redundancy}(b)).

In this work, we fundamentally rethink redundancy in VAR inference by arguing that a truly redundant token or operation is one that has a negligible impact on the pixel domain. We define this impact as latent discrepancy, which measures the necessity of computation by quantifying how much the model’s internal states change during generation. Through systematic analysis, we derive two key findings. (1) As shown in Figure~\ref{Spatial Redundancy}(b), under the guidance of image-latent- or pixel-domain signals, the selected pruning masks become more accurate and yield better performance. (2) In CFG, the conditional--unconditional discrepancy exhibits an overall convergence trend as the scale increases, while the convergence speed varies across different prompts (see Figure~\ref{Guidance Redundancy}(a)). 

Driven by these insights, we propose \textbf{LD-Pruning} (\textbf{L}atent \textbf{D}iscrepancy \textbf{Pruning}), a training-free acceleration framework that selects tokens based on image-level features. As illustrated in Figure~\ref{Spatial Redundancy}(b), although direct pixel-domain selection yields high-quality tokens, it incurs substantial latency. In contrast, selection in the image latent domain achieves a more favorable trade-off between generation quality and efficiency. To this end, we introduce Latent High-Frequency Energy Pruning (LHEP), which approximates pixel-space refinement discrepancy with a decoding-free latent surrogate. By measuring local directional variations in the latent residual map, LHEP identifies regions that are still likely to induce visible fine-scale changes
after decoding.

To further address the dynamic convergence behavior of CFG, we design a Semantic Adaptive Termination Strategy (\textbf{SATS}). Rather than relying on fixed schedules, SATS dynamically monitors the discrepancy between conditional and unconditional branches and adaptively terminates the unconditional branch once it no longer provides meaningful guidance.
Experiments demonstrate that the proposed LD-Pruning significantly accelerates VAR image generation. In summary, our main contributions are as follows:
\begin{itemize}
  \item We introduce a unified perspective, Latent Discrepancy, to characterize the inference dynamics of VAR, revealing that redundancy manifests as sparse spatial refinement and prompt-dependent guidance convergence. 

  \item We propose \textbf{LD-Pruning}, a training-free acceleration framework incorporating LHEP and SATS. LHEP approximates pixel-space refinement discrepancy with a decoding-free Image Latent Domain surrogate for region pruning, while SATS monitors conditional--unconditional discrepancy for adaptive guidance termination, avoiding inconsistent localization and fixed skipping schedules.
  
  \item Extensive experiments demonstrate that LD-Pruning achieves a superior efficiency--quality trade-off, delivering up to 2.35$\times$ inference speedup with negligible performance degradation on standard benchmarks.
\end{itemize}

\section{Related Work}

\subsection{Autoregressive Visual Generation}
Visual Autoregressive (VAR) models \cite{tian2024visual} depart from the sequential token-by-token prediction of traditional AR models \cite{sun2024autoregressive,yu2022scaling}, introducing a next-scale prediction paradigm. This process generates images by transitioning from coarse layouts to fine details, echoing the refinement patterns common in diffusion models \cite{chen2023attentive,feng2025dit4edit,couairon2022diffedit,ho2022imagen,ho2022video,ma2025follow}. Building on this, Infinity \cite{han2025infinity} implements a bitwise tokenizer and classifier along with a self-correction mechanism to enhance detail accuracy. While Infinity scales components to expand model capacity, the exponential growth of tokens at finer resolutions creates significant computational and memory overhead. Consequently, current research emphasizes minimizing redundant computation during these final refinement stages. 

\subsection{Efficient Visual Generation}
Efficiency in diffusion models has been extensively studied through training-based \cite{salimans2022progressive,feng2024relational,ma2024learning,shen2025lazydit} and training-free methods \cite{wang2024attention,hu2024token,ma2024deepcache,zou2024accelerating}, though these methods do not directly translate to the hierarchical generation of VAR. In the autoregressive domain, researchers have explored various efficient decoding strategies \cite{wang2025simplear,teng2024accelerating,jang2024lantern} to mitigate the latency of high-fidelity synthesis. However, the inherent sequential nature of traditional AR models remains a persistent bottleneck as image resolution and token counts increase. 

Recent research \cite{li2025memory,chen2025frequency,aiello2025dreamcache} has begun to specifically address the computational overhead of the VAR paradigm. FastVAR \cite{guo2025fastvar} applies post-training token pruning with cached token reuse but depends on a fixed pruning ratio. SkipVAR \cite{li2025skipvar} introduces a learnable decision mechanism based on handcrafted frequency features to regulate step-skipping and branch replacement. However, SkipVAR’s acceleration remains coarse-grained as it operates primarily at the scale level, and its unconditional branch replacement strategy is fixed regardless of the dynamic generative state. In contrast, our LD-Pruning uses latent discrepancy to couple acceleration decisions more directly with image-level changes.

\section{Preliminary}
We first summarize the VAR formulation used throughout the paper. VAR redefines AR for images by shifting from next-token prediction to next-scale prediction. In this framework, each autoregressive operation generates a token map at a specific resolution scale rather than predicting individual tokens step by step. Given a continuous image feature map $\mathbf{F} \in \mathbb{R}^{h \times w \times d}$, VAR first quantizes it into $K$ multi-scale token maps $\mathbf{R} = (\mathbf{R}_1, \mathbf{R}_2, \ldots, \mathbf{R}_K)$ with increasingly larger predefined scale $(h_k, w_k)$ for $k = 1, \ldots, K$. This sequence of residuals allows us to reconstruct the continuous feature $\mathbf{F}$ as $\mathbf{F}_k = \sum_{i=1}^{k} \operatorname{Up}(\mathbf{R}_i, (h, w))$, where $\operatorname{Up}(\cdot)$ represents the upsampling operation. The multi-scale token maps $\mathbf{R}$ allow the decomposition of the joint probability distribution in an autoregressive manner:

\begin{figure}[t]
\centering
\includegraphics[width=1.\columnwidth]{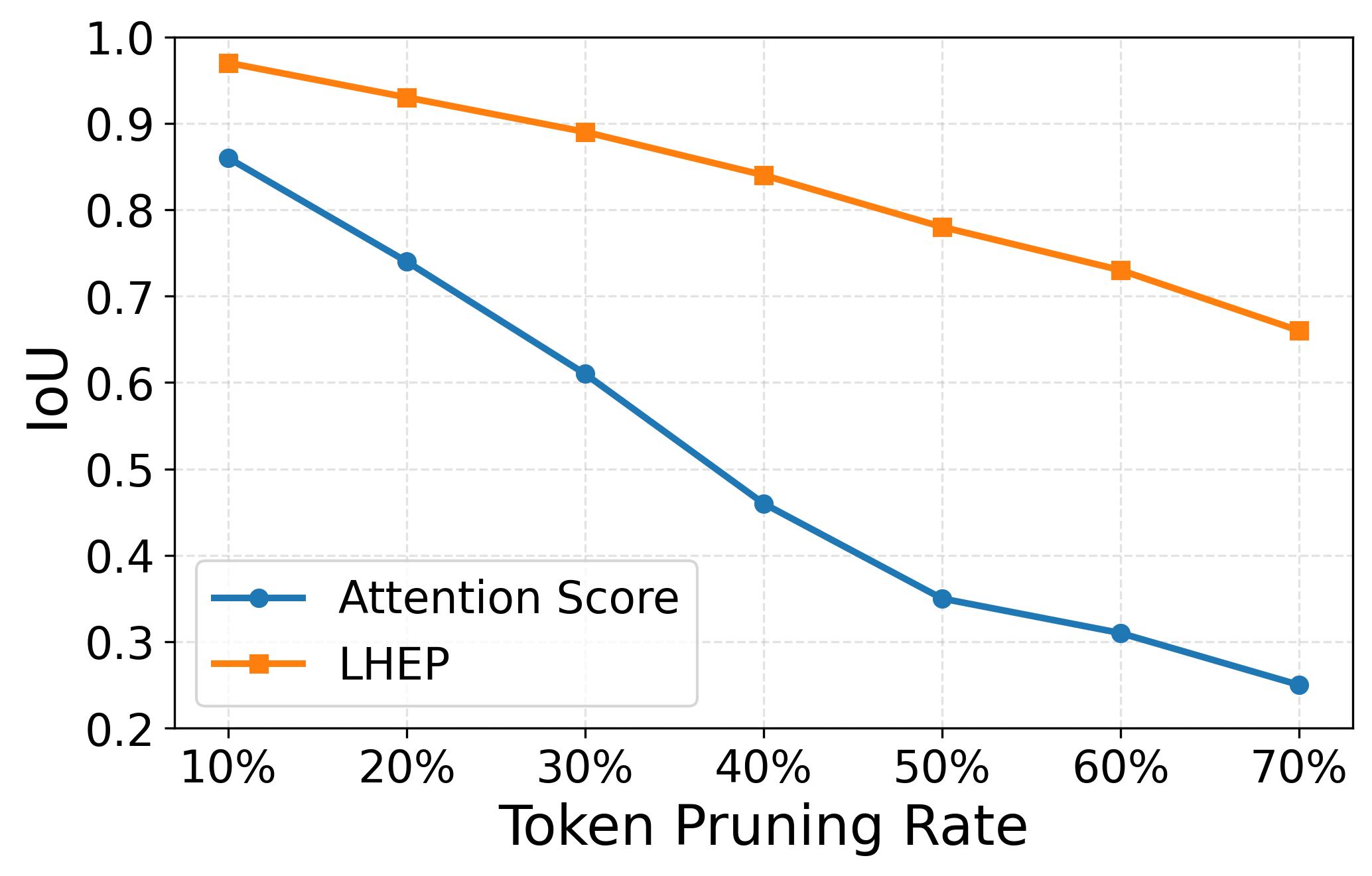}
\caption{
Domain-level alignment with the pixel-space reference signal.
The reference regions are obtained from decoded image differences $|I_k-I_{k-1}|$.
Attention score represents selection in the Semantic Aggregation Domain, while LHEP represents selection in the Image Latent Domain.
}
\label{fig:oracle_iou}
\end{figure}

\begin{equation}
p(\mathbf{R}_1, \mathbf{R}_2, \ldots, \mathbf{R}_K) = \prod_{k=1}^{K} p(\mathbf{R}_k \mid \mathbf{R}_1, \mathbf{R}_2, \ldots, \mathbf{R}_{k-1}),
\end{equation}
where the initial token map $\mathbf{R}_1$ is derived from the text embeddings. For subsequent steps, the input feature $\widetilde{\mathbf{F}}_{k-1}$ is first obtained by downsampling the previous accumulated feature:
\begin{equation}
\widetilde{\mathbf{F}}_{k-1} = \operatorname{Down}(\mathbf{F}_{k-1}, (h_{k-1}, w_{k-1})),
\end{equation}
where $\operatorname{Down}(\cdot)$ denotes the downsampling operation. Then, the VAR transformer $\phi$ predicts output features $\mathbf{F}_k^o$, which are quantized via $\mathcal{Q}$ with codebook $\mathcal{Z} \in \mathbb{R}^{V \times d}$:
\begin{align}
\mathbf{F}_k^o &= \phi(\widetilde{\mathbf{F}}_{k-1}), \\
\mathbf{R}_k &= \mathcal{Q}(\mathbf{F}_k^o).
\end{align}
Here, $\mathbf{F}_k^o \in \mathbb{R}^{M \times d}$ with $M = h_k \times w_k$, and $\mathbf{R}_k$ consists of $h_k \times w_k$ discrete tokens selected from a vocabulary of size $V$ at scale $k$. Finally, the updated feature $\mathbf{F}_k$ is obtained by adding the upsampled residual to the previous feature:
\begin{equation}
\mathbf{F}_k = \mathbf{F}_{k-1} + \operatorname{Up}(\mathbf{R}_k, (h_K, w_K)).
\end{equation}

We initialize $\mathbf{F}_0 = \mathbf{0}$ and $\widetilde{\mathbf{F}}_0 = \langle \text{SOS} \rangle \in \mathbb{R}^{1 \times 1 \times d}$ with $\langle \text{SOS} \rangle$ being the start-of-sequence token. The VAR paradigm generates images in a coarse-to-fine manner with $K$ scale-up steps.

\begin{table}[t]
\centering
\caption{
Comparison of token selection criteria from different domains on Infinity-2B.
}
\label{tab:proxy_comparison}
\resizebox{\columnwidth}{!}{
\begin{tabular}{llccccc}
\toprule
Criterion & Domain & GenEval $\uparrow$ & PSNR $\uparrow$ & SSIM $\uparrow$ & LPIPS $\downarrow$ & Speedup $\uparrow$ \\
\midrule
Attention Score & Semantic Aggregation & 0.7157 & 25.4434 & 0.8343 & 0.1810 & 1.34$\times$ \\
Token Variance & Image Latent & 0.7291 & 26.5492 & 0.8607 & 0.1487 & \textbf{1.73}$\times$ \\
Feature Magnitude & Image Latent & 0.7308 & 27.0651 & 0.8687 & 0.1427 & \textbf{1.73}$\times$ \\
LHEP & Image Latent & \textbf{0.7334} & \textbf{27.3853} & \textbf{0.8769} & \textbf{0.1348} & 1.64$\times$ \\
\bottomrule
\end{tabular}
}
\end{table}

\section{Rethinking Redundancy via Latent Discrepancy}
Having established the VAR formulation, we introduce a unified perspective, \textbf{Latent Discrepancy}, to characterize the inference dynamics. It is defined as the magnitude of information update required by the model's internal states to transition from scale $k-1$ to $k$.Based on latent discrepancy, we systematically analyze redundancy in VAR from a more fundamental perspective.

\subsection{Spatial Redundancy: Where to Refine?}
\label{sec:rethinking spatial redundancy}
Existing acceleration strategies typically rely on  heuristics of the semantic-aggregation domain layer features to identify spatial redundancy. Due to complex contextual relationships, it is difficult to reliably identify truly redundant tokens from semantic-aggregation features alone. As shown in Figure~\ref{Spatial Redundancy}(b), this mismatch leads to inconsistent localization: it may overlook regions that actually require the generation of texture details.

To establish a reliable criterion for ``where to refine,'' we analyze the generative discrepancy observed in the decoded pixel space. We approximate this discrepancy using $|I_k - I_{k-1}|$. As shown in Figure~\ref{Spatial Redundancy}(a), this metric reveals a clear functional decoupling: early scales construct global geometry, while later scales focus predominantly on high-frequency texture refinement (e.g., fur, grass). These pixel-level updates directly reflect the actual accumulation of visual information. Therefore, we use them as a pixel-space reference signal for identifying spatial redundancy. An efficiency-oriented accelerator should aim to bypass regions where this visual update is negligible.

Although the explicit pixel domain can offer precise guidance, accessing it requires full image decoding at every step, which incurs prohibitive computational overhead (see Figure~\ref{Spatial Redundancy}(b)).
Our goal is therefore not to infer token importance from the Semantic Aggregation
Domain, but to estimate the pixel-space refinement discrepancy in the Image Latent
Domain without decoding at every scale.
Ideally, for a local region $\Omega$, a direct pixel-level refinement score can be defined as
\begin{equation}
D_k(\Omega)=\left\|\mathcal{D}(F_k)_\Omega-\mathcal{D}(F_{k-1})_\Omega\right\|,
\label{eq:pixel_refinement_score}
\end{equation}
where $\mathcal{D}$ denotes the image decoder.
A large $D_k(\Omega)$ indicates that the region still produces visible image changes and should be refined, while a small value suggests that the region is likely redundant.
Since computing this score requires decoding at every scale, we approximate it in the latent space.
Because VAR updates the image representation through residual accumulation, late-stage refinements mainly appear as localized residual changes in the latent feature map.
These local high-frequency variations are more likely to induce visible fine-scale changes after decoding.
Therefore, a suitable decoding-free proxy should preserve spatial locality and be sensitive to high-frequency residual variation.
Motivated by this, we use Latent High-frequency Energy as a lightweight proxy for pixel-space refinement discrepancy.
Unlike semantic-aggregation heuristics, it is computed in the Image Latent Domain and directly measures local activeness of undecoded features, enabling efficient localization of regions that still require texture refinement.

We further quantify how well different domain signals align with the pixel-space reference signal. Using decoded image differences $|I_k-I_{k-1}|$ as reference refinement regions, Figure~\ref{fig:oracle_iou} reports the IoU between the
reference and selected regions under different pruning rates. LHEP shows higher overlap than attention-score selection, suggesting that Image Latent Domain cues better preserve the localization of refinement-critical regions than Semantic Aggregation Domain cues. This domain-level trend is also reflected in Table~\ref{tab:proxy_comparison}, where Image Latent Domain criteria consistently improve generation fidelity over the Semantic Aggregation Domain baseline, and LHEP achieves better overall quality while maintaining comparable efficiency. These observations support latent high-frequency energy as a decoding-free proxy for pixel-space refinement discrepancy.

\noindent\textbf{Insight 1:}
Spatial redundancy in fine-scale VAR generation should be identified by
pixel-space refinement discrepancy rather than by token-importance cues from the
Semantic Aggregation Domain.
Since direct pixel-space measurement is expensive, latent high-frequency energy
serves as a decoding-free Image Latent Domain surrogate that preserves locality
and captures directional residual variations for fine visual refinement.

\subsection{Guidance Redundancy: When to Stop?}
High-quality visual generation typically requires computing an unconditional branch alongside the text-conditional one, effectively doubling the per-step computation. Existing acceleration methods address this overhead using static skipping schedules, such as terminating the unconditional branch after a fixed number of steps. However, this uniform strategy overlooks semantic diversity across inputs. As shown in Figure~\ref{Guidance Redundancy}(a) right, static schedules lead to a suboptimal trade-off: early termination degrades semantically complex samples, while prolonged guidance wastes computation on simpler ones.

To determine an appropriate termination scale, we define guidance discrepancy as the difference between the conditional and unconditional branches. As shown in Figure~\ref{Guidance Redundancy}(a) left, this discrepancy is not strictly monotonic across scales, but shows a clear overall downward trend and stays close to zero after early coarse scales, with only small fluctuations caused by local refinements. We further test its impact on generation quality by terminating the unconditional branch from different scales. Figure~\ref{Guidance Redundancy}(a) right shows that simple prompts reach stable SSIM earlier, whereas complex prompts continue to benefit from guidance until later scales. These results indicate that guidance redundancy is semantically dependent rather than temporally fixed. Therefore, effective acceleration requires adaptive discrepancy monitoring instead of fixed skipping schedules.

\noindent\textbf{Insight 2:} Guidance redundancy is semantically dependent rather than temporally fixed. The conditional--unconditional discrepancy exhibits an overall convergence trend,
but it is not required to be strictly monotonic at every scale. The convergence state of this discrepancy provides a reliable real-time signal for adaptively terminating the unconditional branch once it provides negligible additional guidance.

\begin{figure*}[t]
  \begin{center}
    \centerline{\includegraphics[width=0.98\textwidth]{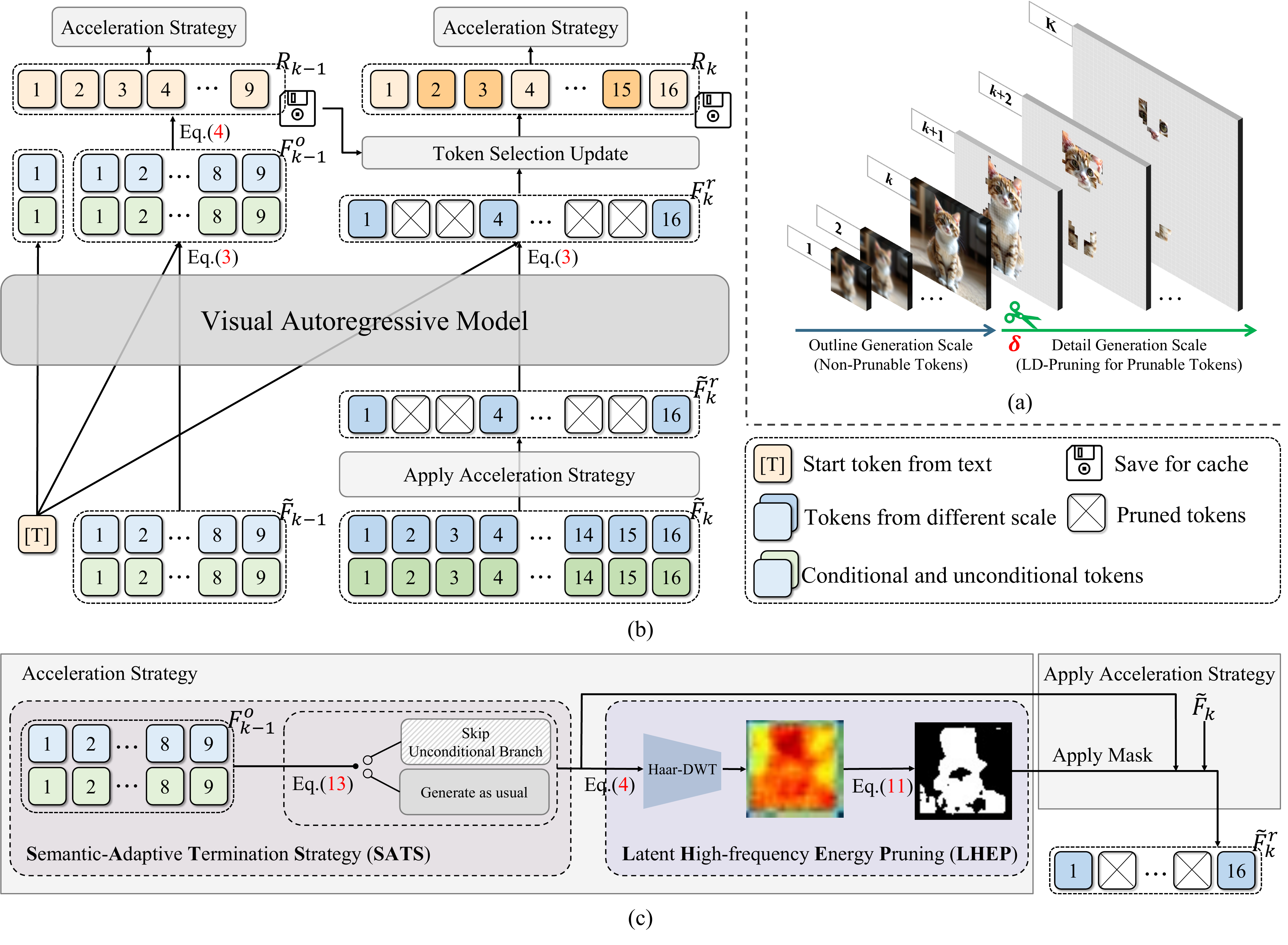}}
    \caption{(a) Overall pipeline of the proposed LD-Pruning. (b) Overview of the proposed LD-Pruning framework. We gather active tokens for efficient sparse inference, then scatter the computed residuals back into the upsampled base map to reconstruct the full spatial state. (c) \textbf{Core Mechanisms.} \textbf{SATS} (left) monitors real-time guidance discrepancy to adaptively terminate the unconditional branch when convergence is detected. \textbf{LHEP} (middle) leverages Haar-DWT to extract latent high-frequency energy, generating a dynamic mask to precisely filter out spatially redundant tokens.
    }
    \label{pipeline}
  \end{center}
  \vskip -0.2in
\end{figure*}

\section{Method}
Based on the insights above, we propose \textbf{LD-Pruning}, a training-free acceleration framework. As illustrated in Figure~\ref{pipeline}, it incorporates \textbf{LHEP} to eliminate spatial redundancy and \textbf{SATS} to remove guidance redundancy.

\subsection{Latent High-frequency Energy Pruning (LHEP)}
To effectively eliminate spatial redundancy without the computational overhead of pixel-space decoding, LHEP estimates pixel-space refinement discrepancy in the Image Latent Domain by measuring local high-frequency residual variations. This process involves three key steps: Haar wavelet decomposition, local energy aggregation, and distribution-adaptive selection.

\paragraph{Latent Haar Wavelet Decomposition.}
To capture the \textit{Latent Discrepancy} defined in Section~\ref{sec:rethinking spatial redundancy}, we focus on the quantized token map $\mathbf{R}_{k-1} \in \mathbb{R}^{C \times {h}_{k-1} \times {w}_{k-1}}$ from scale $k-1$.
Since VAR generates images through residual accumulation, $\mathbf{R}_{k-1}$ naturally reflects the incremental update from the previous scale.
Regions with strong local residual variations are more likely to induce visible fine-scale refinements after decoding, whereas regions with weak variations are more likely to have stabilized.
Therefore, measuring local high-frequency residual energy provides a decoding-free estimate of where refinement is still necessary.

To extract such local residual variations efficiently, we employ the 2D discrete Haar Wavelet Transform (Haar-DWT).
Haar-DWT provides a minimal local finite-difference basis for measuring residual variations within each latent patch.
For each $2 \times 2$ latent patch, the Haar basis decomposes the signal into one low-frequency component and three directional high-frequency components.
Since our objective is region selection rather than full signal reconstruction, the low-frequency component mainly reflects the local average state, while the high-frequency components measure whether the patch still contains directional residual changes.
Moreover, Haar-DWT is static and parameter-free, avoiding additional inference branches or learned predictors.
This approach utilizes four fixed kernels to disentangle the low-frequency approximation ($\mathbf{K}_{LL}$) from directional details ($\mathbf{K}_{LH}, \mathbf{K}_{HL}, \mathbf{K}_{HH}$):
\begin{equation}
\small
\begin{aligned}
    \mathbf{K}_{LL} &= \frac{1}{2}\begin{bmatrix} 1 & 1 \\ 1 & 1 \end{bmatrix}, & 
    \mathbf{K}_{LH} &= \frac{1}{2}\begin{bmatrix} -1 & 1 \\ -1 & 1 \end{bmatrix}, \\
    \mathbf{K}_{HL} &= \frac{1}{2}\begin{bmatrix} -1 & -1 \\ 1 & 1 \end{bmatrix}, & 
    \mathbf{K}_{HH} &= \frac{1}{2}\begin{bmatrix} -1 & 1 \\ 1 & -1 \end{bmatrix}.
\end{aligned}
\end{equation}
These high-frequency kernels explicitly capture horizontal, vertical, and diagonal residual variations within the latent feature map.
We compute the raw high-frequency energy map $\mathbf{E}_{raw} \in \mathbb{R}^{\frac{h_{k-1}}{2} \times \frac{w_{k-1}}{2}}$ by aggregating the squared high-frequency responses across all channels:
\begin{equation}
    \mathbf{E}_{raw} = \sum_{d \in \{LH, HL, HH\}} \| \mathbf{R}_{k-1} * \mathbf{K}_d \|^2_2.
    \label{eq:raw_energy}
\end{equation}

\paragraph{Local Energy Aggregation.} 
Raw high-frequency signals can be spatially fragmented. To ensure the spatial continuity of the selected regions and incorporate neighborhood dependencies, we apply a local smoothing operation using Average Pooling ($3 \times 3$ kernel, stride=1, padding=1), followed by interpolation to match the target resolution:
\begin{equation}
    \mathbf{E}_{target} = \operatorname{Interpolate}(\operatorname{AvgPool2d}(\mathbf{E}_{raw})).
\end{equation}
This step provides a coherent indication of generative necessity, preventing isolated noise from triggering redundant computations.

\paragraph{Distribution-Adaptive Selection.} 
A fixed pruning threshold is suboptimal due to the varying texture complexity across images. Instead, we propose a dynamic selection criterion based on the Cumulative Distribution Function (CDF) of the energy. We flatten $\mathbf{E}_{target}$ into a vector $\mathbf{v}$ and sort it in descending order. We then determine the dynamic threshold $\tau$ that preserves the target energy ratio $\rho$:
\begin{equation}
\small
    k^* = \min \left\{ k \mid \sum_{i=1}^k \mathbf{v}_{sorted}^{(i)} \ge \rho \|\mathbf{v}\|_1 \right\}, \quad \tau = \mathbf{v}_{sorted}^{(k^*)}.
\end{equation}
Based on $\tau$, we generate the binary pruning mask $\mathbf{M}$ element-wise:
\begin{equation}
    \mathbf{M}^{(i,j)} = 
    \begin{cases} 
    1, & \text{if } \mathbf{E}_{target}^{(i,j)} \geq \tau \\ 
    0, & \text{otherwise.}
    \end{cases}
\end{equation}
This mask explicitly guides the sparse calculation for the next scale $\tilde{F}_k$: only tokens corresponding to active regions ($\mathbf{M}^{(i,j)}=1$) are computed, whereas the remaining tokens are skipped to reduce redundancy (See Figure~\ref{pipeline}(c) middle).

\subsection{Semantic-Adaptive Termination Strategy (SATS)}
To address guidance redundancy, SATS dynamically terminates the unconditional branch based on the real-time monitoring of guidance discrepancy.

\paragraph{Guidance Convergence Monitoring.} 
Instead of relying on static steps, we monitor the evolution of the guidance magnitude to determine its necessity. At each scale $k$, we first compute the absolute guidance magnitude $\Delta_k$ as the $L_2$-norm distance between the conditional logits $l_{cond}^{(k)}$ and unconditional logits $l_{uncond}^{(k)}$:
\begin{equation}
    \Delta_k = \| l_{cond}^{(k)} - l_{uncond}^{(k)} \|_2.
\end{equation}
To capture the dynamic behavior of this guidance, we compute the \textit{relative change rate} $\gamma_k$ between adjacent scales: 
\begin{equation}
    \gamma_k = \frac{| \Delta_k - \Delta_{k-1} |}{\Delta_{k-1} + \epsilon},
    \label{eq:relative_change}
\end{equation}
where $\epsilon$ is a small constant for numerical stability. This metric $\gamma_k$ serves as a real-time indicator of guidance convergence (See Figure~\ref{pipeline}(c) left).

\paragraph{Adaptive Termination.} 
We define a convergence tolerance $\delta$. Upon satisfying the condition $\gamma_k < \delta$, we terminate the computation of the unconditional branch for all subsequent scales, effectively switching the model to single-branch inference to save computation.

\begin{table*}[t]
\caption{\textbf{Quantitative comparison} on GenEval and DPG-Bench. Latency is measured on a single GPU with batch size 1.}
\label{GenEval and DPG}
\centering
\small
\renewcommand{\arraystretch}{0.9}
\setlength{\tabcolsep}{5.5pt}
\begin{tabular}{lccccccccccc}
\toprule
\multirow{2}{*}{Methods} & \multicolumn{5}{c}{GenEval} & \multicolumn{4}{c}{DPG-Bench} & \multirow{2}{*}{Latency(s)$\downarrow$} & \multirow{2}{*}{Speedup} \\
\cmidrule(lr){2-6} \cmidrule(lr){7-10}
& Two Obj. & Position & Color & Attrs. & Overall & Global & Entity & Relation & Overall & & \\
\midrule
Infinity-2B & 83.59 & 44.50 & 84.57 & 58.25 & 0.74 & 88.50 & 88.12 & 88.77 & 82.84 & 0.90 & 1.00$\times$ \\
+FastVAR & 82.83 & 44.75 & 81.91 & 54.50 & 0.71 & 81.42 & 88.27 & \textbf{90.13} & 82.65 & 0.62 & 1.45$\times$ \\
+SkipVAR & \textbf{83.33} & 44.75 & \textbf{84.31} & 56.00 & \textbf{0.73} & 85.27 & 89.11 & 89.52 & \textbf{82.88} & 0.65 & 1.38$\times$ \\
\rowcolor{gray!20}
+LD-Pruning & 82.07 & \textbf{45.25} & 83.78 & \textbf{57.00} & \textbf{0.73} & \textbf{86.97} & \textbf{89.95} & 90.11 & 82.65 & 0.52 & \textbf{1.73$\times$} \\
\midrule
Infinity-8B & 87.88 & 61.50 & 86.97 & 68.25 & 0.79 & 92.14 & 89.74 & 92.08 & 86.44 & 1.65 & 1.00$\times$ \\
+FastVAR & 86.62 & 60.75 & \textbf{86.97} & \textbf{67.00} & 0.78 & 87.30 & \textbf{91.46} & 91.11 & 86.33 & 0.92 & 1.79$\times$ \\
+SkipVAR & \textbf{87.88} & \textbf{61.00} & 85.11 & 66.75 & 0.78 & \textbf{92.64} & 90.59 & 90.09 & 86.29 & 0.97 & 1.71$\times$ \\
\rowcolor{gray!20}
+LD-Pruning & 87.63 & \textbf{61.00} & 86.44 & \textbf{67.00} & \textbf{0.79} & 91.56 & 89.61 & \textbf{94.45} & \textbf{86.34} & 0.70 & \textbf{2.35$\times$} \\
\midrule
HART & 53.28 & 18.00 & 84.31 & 18.25 & 0.51 & 85.74 & 82.25 & 85.21 & 74.75 & 0.45 & 1.00$\times$ \\
+FastVAR & 48.74 & 16.50 & 81.65 & 16.25 & 0.49 & 79.37 & 82.48 & \textbf{85.89} & 74.76 & 0.40 & 1.13$\times$ \\
\rowcolor{gray!20}
+LD-Pruning & \textbf{53.03} & \textbf{17.75} & \textbf{83.78} & \textbf{16.50} & \textbf{0.50} & \textbf{84.22} & \textbf{83.02} & 85.57 & \textbf{75.61} & 0.34 & \textbf{1.32$\times$} \\
\bottomrule
\end{tabular}
\end{table*}

\begin{table*}[t]
\caption{\textbf{Quantitative comparison} on HPSv2.1, ImageReward, GenEval, PSNR, SSIM and LPIPS. Latency is measured on a single GPU with batch size 1. \textsuperscript{1} denotes latency for preference benchmarks, \textsuperscript{2} denotes latency for generation quality benchmarks.}
\label{combined HPSv-low level}
\centering
\small
\renewcommand{\arraystretch}{0.9}
\setlength{\tabcolsep}{5pt}
% \begin{tabular}{@{}lcccccccccccc@{}}
\begin{tabular}{@{}lcccccc@{\hspace{6pt}\vrule\hspace{6pt}}cccccc@{}}
\toprule
\multirow{2}{*}{Methods} & \multicolumn{3}{c}{HPSv2.1} & \multirow{2}{*}{\makecell{ImgRw.$\uparrow$}} & \multirow{2}{*}{\makecell{Lat.(s)\textsuperscript{1}$\downarrow$}} & \multirow{2}{*}{\makecell{Spd.\textsuperscript{1}}} & \multirow{2}{*}{\makecell{GenEv.}} & \multirow{2}{*}{\makecell{PSNR$\uparrow$}} & \multirow{2}{*}{\makecell{SSIM$\uparrow$}} & \multirow{2}{*}{\makecell{LPIPS$\downarrow$}} & \multirow{2}{*}{\makecell{Lat.(s)\textsuperscript{2}$\downarrow$}} & \multirow{2}{*}{\makecell{Spd.\textsuperscript{2}}} \\
\cmidrule(lr){2-4}
& Photo & C-Art & Overall & & & & & & & & & \\
\midrule
Infinity-2B & 29.45 & 30.47 & 30.55 & 0.94 & 0.72 & 1.00$\times$ & 0.74 & -- & -- & -- & 0.90 & 1.00$\times$ \\
+FastVAR & 28.80 & 29.91 & 29.97 & \textbf{0.92} & 0.53 & 1.36$\times$ & 0.71 & 22.26 & 0.7787 & 0.2424 & 0.62 & 1.45$\times$ \\
\rowcolor{gray!20}
+LD-Pruning & \textbf{29.29} & \textbf{30.30} & \textbf{30.39} & \textbf{0.92} & 0.51 & \textbf{1.41$\times$} & \textbf{0.73} & \textbf{24.85} & \textbf{0.8239} & \textbf{0.1842} & 0.52 & \textbf{1.73$\times$} \\
\cdashline{1-13}
\\
[-0.8em]
+SkipVAR & 29.31 & 30.38 & \textbf{30.47} & \textbf{0.94} & 0.57 & 1.26$\times$ & \textbf{0.73} & 26.47 & 0.8768 & 0.1416 & 0.65 & 1.38$\times$ \\
\rowcolor{gray!20}
+LD-Pruning & \textbf{29.35} & \textbf{30.39} & \textbf{30.47} & \textbf{0.94} & 0.55 & \textbf{1.31$\times$} & \textbf{0.73} & \textbf{27.42} & \textbf{0.8785} & \textbf{0.1350} & 0.64 & \textbf{1.41$\times$} \\
\midrule
Infinity-8B & 29.49 & 31.28 & 31.00 & 1.05 & 1.47 & 1.00$\times$ & 0.79 & -- & -- & -- & 1.65 & 1.00$\times$ \\
+FastVAR & 28.72 & 30.26 & 30.09 & 1.02 & 0.76 & 1.93$\times$ & 0.78 & 22.66 & 0.7238 & 0.2166 & 0.92 & 1.79$\times$ \\
\rowcolor{gray!20}
+LD-Pruning & \textbf{28.92} & \textbf{30.45} & \textbf{30.24}  & \textbf{1.03} & 0.69 & \textbf{2.13$\times$} & \textbf{0.79} & \textbf{23.82} & \textbf{0.7644} & \textbf{0.1726} & 0.70 & \textbf{2.35$\times$} \\
\cdashline{1-13}
\\
[-0.8em]
+SkipVAR & 29.09 & 30.90 & 30.64 & \textbf{1.03} & 0.92 & 1.60$\times$ & \textbf{0.79} & 25.86 & 0.8540 & 0.1243 & 0.97 & 1.71$\times$ \\
\rowcolor{gray!20}
+LD-Pruning & \textbf{29.21} & \textbf{30.93} & \textbf{30.65} & \textbf{1.03} & 0.86 & \textbf{1.71$\times$} & \textbf{0.79} & \textbf{26.87} & \textbf{0.8622} & \textbf{0.1135} & 0.95 & \textbf{1.73$\times$} \\
\midrule
HART & 27.62 & 29.00 & 29.07 & 0.66 & 0.35 & 1.00$\times$ & 0.51 & -- & -- & -- & 0.45 & 1.00$\times$ \\
+FastVAR & 26.26 & 27.66 & 27.68 & 0.60 & 0.30 & 1.17$\times$ & 0.49 & 20.85 & 0.7044 & 0.3161 & 0.40 & 1.13$\times$ \\
\rowcolor{gray!20}
+LD-Pruning & \textbf{27.13} & \textbf{28.76} & \textbf{28.84}  & \textbf{0.64} & 0.27 & \textbf{1.30$\times$} & \textbf{0.50} & \textbf{22.15} & \textbf{0.7454} & \textbf{0.2325} & 0.34 & \textbf{1.32$\times$} \\
\bottomrule
\end{tabular}
\vskip -0.15in
\end{table*}

\subsection{Token Selection Update}
Upon determining the pruning mask $\mathbf{M}$, we gather the $r = \sum \mathbf{M}^{(i,j)}$ active inputs into a sparse tensor $\tilde{F}_{k}^{r}$. This compact representation undergoes sparse inference via the VAR transformer to predict output features $\mathbf{F}_k^r = \phi(\tilde{F}_{k}^{r})$, which are subsequently discretized into sparse residual tokens $\mathbf{R}_k^r = \mathcal{Q}(\mathbf{F}_k^r)$ via quantization. This selective computation significantly reduces redundancy by processing only the regions requiring refinement.

To reconstruct the full spatial token map $\mathbf{R}_k$ for the subsequent scale, we approximate the skipped regions by upsampling the previous map $\mathbf{R}_{base} = \text{Up}(\mathbf{R}_{k-1}, (h_k, w_k))$ using nearest-neighbor interpolation. As shown in Figure~\ref{pipeline}(b), the final state update is achieved by scattering the computed sparse residuals $\mathbf{R}_k^r$ into this base map at active positions, ensuring both newly generated details and inherited global structures are preserved for the next iteration:
\begin{equation}
    \mathbf{R}_k^{(i,j)} = 
    \begin{cases} 
        \mathbf{R}_k^r, & \text{if } \mathbf{M}^{(i,j)} = 1 \\
        \mathbf{R}_{base}^{(i,j)}, & \text{otherwise.}
    \end{cases}
\end{equation}

\begin{figure*}[t]
  % \vskip 0.2in
  \begin{center}
    \centerline{\includegraphics[width=1.0\textwidth]{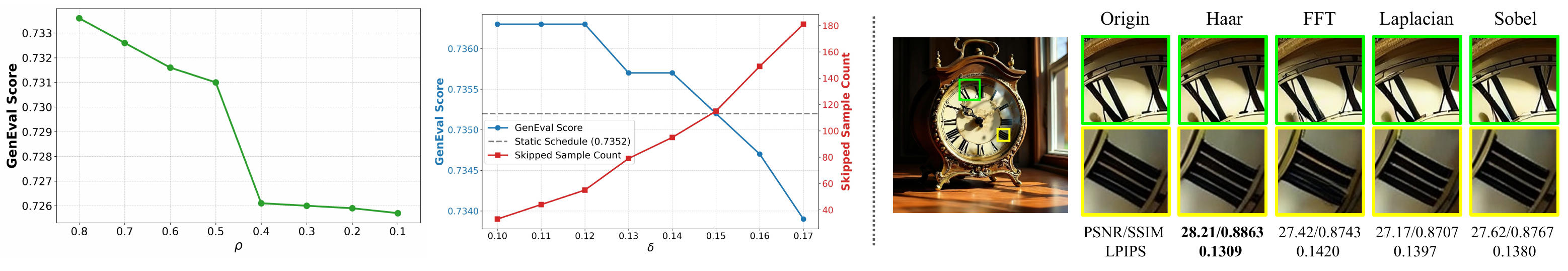}}
    \caption{
      \textbf{Left:} Ablation on LHEP threshold $\rho$ and SATS threshold $\delta$. \textbf{Right:} Ablation on high-frequency extraction operators.
    }
    \label{Ablation study}
  \end{center}
  \vskip -0.2in
\end{figure*}

\begin{figure*}[t]
\begin{center}\centerline{\includegraphics[width=1.\textwidth]{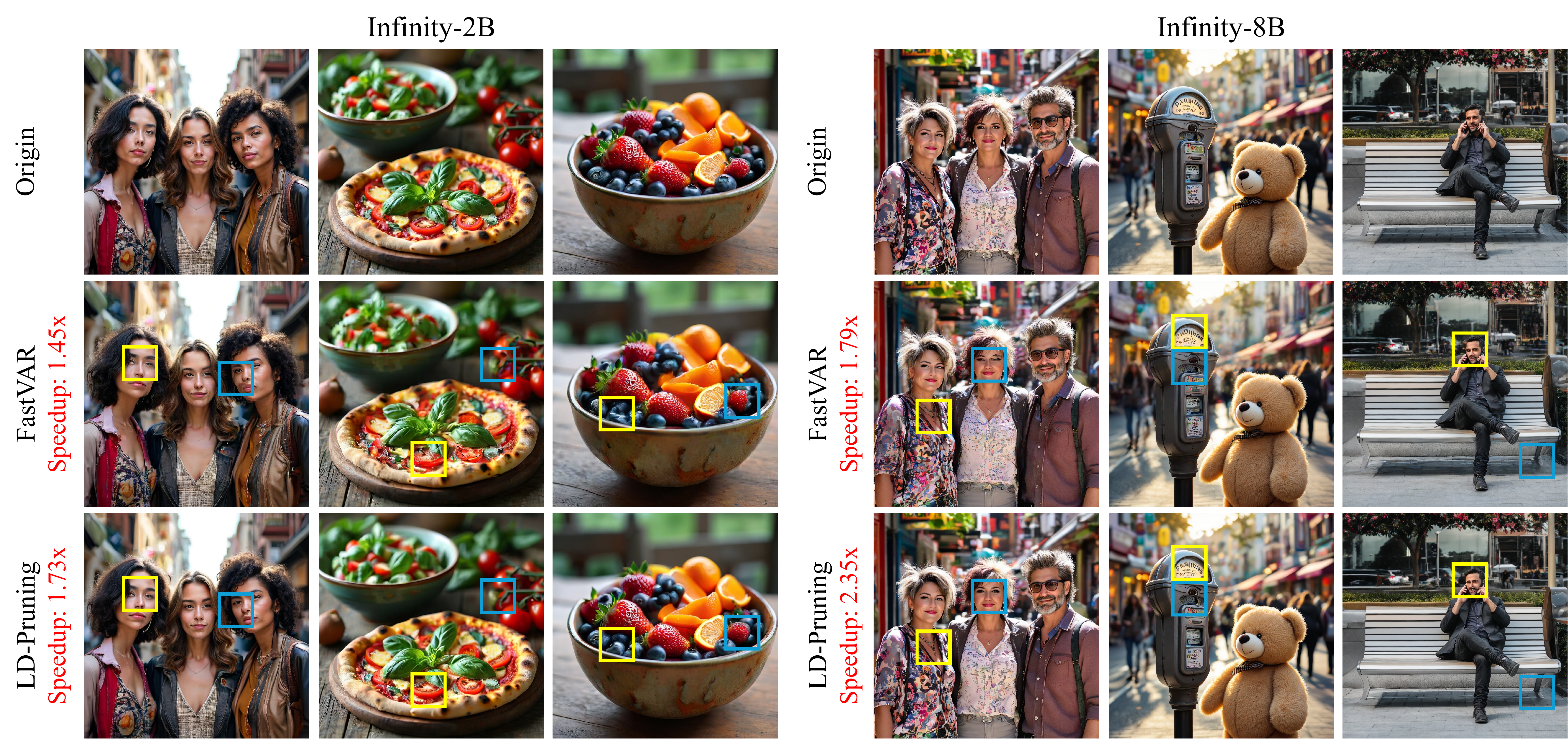}}
    \caption{\textbf{Qualitative comparison} of various methods.
    }
    \label{visual}
  \end{center}
  \vskip -0.4in
\end{figure*}

\begin{table}[t]
\caption{\textbf{Ablation study} of LHEP and SATS on Infinity-2B in terms of GenEval and efficiency.}
\label{LHEP-SATS}
\centering
\small
\begin{tabular}{lccc}
\toprule
Method & Latency(s)$\downarrow$ & Speed$\uparrow$ & GenEval$\uparrow$ \\
\midrule
Infinity-2B & 0.90 & $1.00\times$ & 0.7385 \\
++ LHEP & 0.55 & $1.64\times$ & 0.7334 \\
++ SATS & 0.73 & $1.23\times$ & 0.7352 \\
++ LHEP \& SATS & 0.52 & $\mathbf{1.73}\times$ & 0.7302 \\
\bottomrule
\end{tabular}
\vskip -0.2in
\end{table}

\section{Experiments}
\subsection{Experimental Setup}
\paragraph{Base Models.} We apply our proposed \textbf{LD-Pruning} to representative VAR-based text-to-image baselines, \textbf{Infinity-2B}, \textbf{Infinity-8B}, and \textbf{HART}, to validate its generality across model scales. For a fair comparison, we follow the official Infinity and HART implementations for all hyperparameters and configurations. We keep the original inference for scales 0 to 12 and accelerate only the detail-generation scales (scales $\{9,10,11,12\}$): \textbf{SATS} at scales 9 to 12 and \textbf{LHEP} at scales 10 to 12. When comparing with \textbf{FastVAR}, we use the same setting by skipping scales 11 to 12 for a fair efficiency comparison. Infinity baselines are evaluated with FlashAttention, and all experiments are run on a single NVIDIA H100 GPU. For reproducibility, we use the same fixed random seed for all methods in all experiments.

\paragraph{Evaluation Metrics.} We evaluate the proposed method from two perspectives: generation quality and inference efficiency. For generation quality, we consider both high-level and low-level. High-level evaluation focuses on semantic alignment and human preference, measured using four popular benchmark--—GenEval \cite{ghosh2023geneval}, DPG-Bench \cite{hu2024ella}, ImageReward \cite{xu2023imagereward} and HPSv2.1 \cite{wu2023human}. For low-level evaluation, we take the baseline model’s outputs as the reference and employ PSNR, SSIM \cite{wang2004image}, and LPIPS \cite{zhang2018unreasonable} to quantitatively assess the preservation of high-frequency textures and visual fidelity.

\subsection{Main Results}
\paragraph{Quantitative Comparison on GenEval and DPG.}
Table \ref{GenEval and DPG} shows that LD-Pruning consistently offers the best speed--quality trade-off across Infinity (2B/8B) and HART. On Infinity-8B, it cuts latency from 1.65s to 0.70s ($2.35\times$) while keeping GenEval Overall $=0.79$ and DPG Overall $=86.34$, both nearly unchanged from the baseline. On HART, LD-Pruning similarly delivers a clear speedup ($1.32\times$) with competitive GenEval/DPG.

\paragraph{Quantitative Comparison on HPSv2 and ImageReward.}
Table \ref{combined HPSv-low level} confirms that LD-Pruning preserves preference-aligned quality on both Infinity and HART. For Infinity-8B, LD-Pruning achieves a $2.13\times$ speedup (1.47s $\rightarrow$ 0.69s) while keeping HPSv2.1 Overall and ImageReward Overall close to the baseline. On HART, LD-Pruning also maintains strong preference scores with meaningful latency reduction.

\paragraph{Quantitative Comparison on Visual Fidelity.}
Table \ref{combined HPSv-low level} further evaluates LD-Pruning better preserves fine details, as reflected by higher PSNR/SSIM and lower LPIPS compared with attention-based pruning. For example, on Infinity-2B, LD-Pruning improves PSNR/SSIM from 22.26/0.7787 (FastVAR) to 27.42/0.8785, and reduces LPIPS from 0.2424 to 0.1350, indicating substantially closer outputs to the baseline. Besides, Figure~\ref{visual} visually shows that LD-Pruning retains high-frequency textures in complex regions where heuristic baselines often blur or lose fine structures. More visual fidelity comparisons are in Appendix \ref{sec:appendix_visual}.

\subsection{Ablation Study}
\paragraph{Ablation of LHEP and SATS}
Table \ref{LHEP-SATS} indicates that LHEP introduces only \textbf{~0.4ms} overhead per execution, which is negligible in practice. Moreover, adding SATS on top of LHEP further reduces latency while keeping GenEval essentially stable.

\paragraph{Ablation Studies on High-frequency Extraction Operators.}
To identify the optimal energy extractor for LHEP, we compare Haar-DWT with FFT, Laplacian, and Sobel operators. As shown in Figure \ref{Ablation study} (Right), Haar-DWT yields the superior reconstruction quality, achieving a PSNR of 28.21 and SSIM of 0.8863. We attribute this success to the Haar-DWT's inherent capacity to capture local directional gradients precisely with zero parameter overhead.

\begin{table}[t]
  \caption{\textbf{Consumer-GPU speed evaluation} on Infinity-2B.}
  \label{3090_speed}
  \centering
        \begin{tabular}{lcccr}
          \toprule
          Methods  & GenEval         & Latency(s)$\downarrow$      & Speedup$\uparrow$   \\
          \midrule
          Infinity-2B    & 0.73 & 2.43& 1.00$\times$  \\
          FastVAR & 0.72 & 0.96 & 2.53$\times$ \\
          SkipVAR    & 0.72 & 1.21 & 2.01$\times$  \\
          LD-Pruning    & 0.72 & 0.81 & \textbf{3.00$\times$}         \\
          \bottomrule
        \end{tabular}
  \vskip -0.2in
\end{table}

\paragraph{Sensitivity Analysis of LHEP Threshold ($\rho$).}
We investigate the energy preservation ratio $\rho$, which governs the sparsity of our spatial mask. As plotted in Figure \ref{Ablation study} (Left), the GenEval remains stable ($\approx 0.73$) for $\rho \ge 0.5$, suggesting that the pruned high-frequency components have limited impact on overall generation quality. When $\rho < 0.4$, quality drops sharply (down to $\approx 0.720$), so we set $\rho = 0.5$ as a reliable operating point. More hyperparameter sensitivity results on Infinity-8B and HART are provided in Appendix~\ref{app:hyper_sensitivity}.

\paragraph{Sensitivity Analysis of SATS Threshold ($\delta$).}
We further analyze the convergence tolerance $\delta$ in SATS. Figure~\ref{Ablation study} (Left) shows that increasing $\delta$ makes SATS skip more unconditional branches, improving efficiency, while excessive skipping harms quality. We choose $\delta = 0.15$, which skips the unconditional branch for about 115 samples while keeping GenEval comparable to the conservative setting.

\paragraph{Consumer-GPU Speed Evaluation (RTX 3090).}
Beyond the H100 profiling in the main experiments, we conduct an additional speed test on a consumer-grade NVIDIA RTX 3090 to evaluate the practical acceleration benefits of LD-Pruning.
As reported in Table~\ref{3090_speed}, LD-Pruning achieves a \textbf{$3.00\times$} speedup over Infinity-2B (2.43s $\rightarrow$ 0.81s) while keeping GenEval essentially unchanged (0.73 $\rightarrow$ 0.72). 
This suggests it remains effective on commodity GPUs, offering a strong efficiency--quality trade-off for practical deployment.

\section{Conclusion}
In this paper, we rethink redundancy in VAR generation from its impact on image-level outputs and introduce \textbf{Latent Discrepancy} to quantify token contribution via state changes.
Our analysis shows that redundancy is more reliably characterized by its \emph{image-level impact} rather than heuristic measures based on layer features. The CFG branch discrepancy converges at a semantic-dependent rate, motivating adaptive unconditional-branch skipping.
Based on these findings, we propose \textbf{LD-Pruning}, a training-free framework with two components: \textbf{LHEP} for decoding-free region selection using latent high-frequency energy, and \textbf{SATS} for adaptive unconditional-branch skipping.
Experiments on Infinity-2B/8B and HART show that LD-Pruning substantially reduces latency while preserving generation quality, achieving up to \textbf{2.35$\times$} speedup on Infinity-8B with negligible degradation on standard benchmarks.

\nocite{langley00}

\bibliography{example_paper}
\bibliographystyle{icml2026}

%%%%%%%%%%%%%%%%%%%%%%%%%%%%%%%%%%%%%%%%%%%%%%%%%%%%%%%%%%%%%%%%%%%%%%%%%%%%%%%
%%%%%%%%%%%%%%%%%%%%%%%%%%%%%%%%%%%%%%%%%%%%%%%%%%%%%%%%%%%%%%%%%%%%%%%%%%%%%%%
% APPENDIX
%%%%%%%%%%%%%%%%%%%%%%%%%%%%%%%%%%%%%%%%%%%%%%%%%%%%%%%%%%%%%%%%%%%%%%%%%%%%%%%
%%%%%%%%%%%%%%%%%%%%%%%%%%%%%%%%%%%%%%%%%%%%%%%%%%%%%%%%%%%%%%%%%%%%%%%%%%%%%%%
\newpage
\appendix
\onecolumn

\section{Derivation of LHEP as a Decoding-free Approximation of Pixel-space Refinement}
\label{app:latent_discrepancy_derivation}

We further elaborate on the pixel-level refinement score introduced in Eq.~\eqref{eq:pixel_refinement_score}, and show how it leads to a decoding-free latent approximation. Let $\mathcal{P}_{\Omega}$ denote the spatial restriction operator that extracts a local image region $\Omega$. In the pixel domain, the direct refinement magnitude between two consecutive scales can be measured as
\begin{equation}
D_k(\Omega)
=
\left\|
\mathcal{P}_{\Omega}
\left(
\mathcal{D}(F_k)-\mathcal{D}(F_{k-1})
\right)
\right\|_2,
\label{eq:app_pixel_refinement}
\end{equation}
where $\mathcal{D}$ is the image decoder. According to the residual accumulation
rule of VAR, the latent state satisfies
\begin{equation}
F_k = F_{k-1} + U_k,
\qquad
U_k = \mathrm{Up}(R_k),
\label{eq:app_residual_update}
\end{equation}
where $U_k$ denotes the upsampled residual update at scale $k$. Therefore, the
decoded refinement can be written as
\begin{equation}
\mathcal{D}(F_k)-\mathcal{D}(F_{k-1})
=
\mathcal{D}(F_{k-1}+U_k)-\mathcal{D}(F_{k-1}).
\label{eq:app_decoded_update}
\end{equation}

Assume that the decoder is locally differentiable around $F_{k-1}$. Applying a
first-order Taylor expansion yields
\begin{equation}
\mathcal{D}(F_{k-1}+U_k)-\mathcal{D}(F_{k-1})
=
J_{\mathcal{D}}(F_{k-1}) U_k
+
\mathcal{R}_k,
\label{eq:app_taylor}
\end{equation}
where $J_{\mathcal{D}}(F_{k-1})$ is the decoder Jacobian and $\mathcal{R}_k$
denotes the higher-order remainder. If the decoder Jacobian is locally Lipschitz
with constant $L_{\mathcal{D}}$, the remainder is bounded by
\begin{equation}
\left\|\mathcal{R}_k\right\|_2
\le
\frac{L_{\mathcal{D}}}{2}\left\|U_k\right\|_2^2.
\label{eq:app_remainder_bound}
\end{equation}
In the late detail-generation stages, $U_k$ mainly represents local refinement
updates and is typically small compared with the accumulated latent state.
Consequently, the first-order term dominates the decoded image change, and
Eq.~\eqref{eq:app_pixel_refinement} can be approximated as
\begin{equation}
D_k(\Omega)
\approx
\left\|
\mathcal{P}_{\Omega}
J_{\mathcal{D}}(F_{k-1}) U_k
\right\|_2.
\label{eq:app_first_order_score}
\end{equation}

For convolutional or locally structured decoders, the decoded response in
$\Omega$ is primarily affected by latent updates in a finite neighborhood
$\mathcal{N}(\Omega)$ around the corresponding latent region. Thus, the
pixel-space refinement score is mainly controlled by local latent residual
changes:
\begin{equation}
D_k(\Omega)
\lesssim
C_{\Omega}
\left\|
\mathcal{P}_{\mathcal{N}(\Omega)} U_k
\right\|_2
+
\mathcal{O}\left(\|U_k\|_2^2\right),
\label{eq:app_local_bound}
\end{equation}
where $C_{\Omega}$ depends on the local decoder sensitivity. This relation does
not imply that latent residual magnitude is identical to pixel-space refinement;
rather, it indicates that local latent residual variation provides an informative
decoding-free signal for estimating where visible refinement may occur.

However, the local residual magnitude in Eq.~\eqref{eq:app_local_bound} contains
both low-frequency and high-frequency components. In late VAR scales, the global
layout and coarse color structure have largely been established, while visible
changes are dominated by local texture, boundary, and detail refinements.
Therefore, for region selection, the high-frequency component of the local
residual is more informative than the local average component. Let
$\mathcal{H}$ denote a local high-pass operator. We use the following
decoding-free latent refinement score:
\begin{equation}
\widehat{D}_k(\Omega)
=
\left\|
\mathcal{H}
\mathcal{P}_{\mathcal{N}(\Omega)}
U_k
\right\|_2^2.
\label{eq:app_latent_hf_score}
\end{equation}

In practice, when selecting tokens for the next scale, the future residual $U_k$ is not available before computing scale $k$. LHEP therefore uses the latest available residual token map $R_{k-1}$ as a predictive signal of the local refinement state. The practical score is written as
\begin{equation}
\widehat{D}_{k-1}(\Omega)
=
\sum_{c=1}^{C}
\left\|
\mathcal{H}
\mathcal{P}_{\mathcal{N}(\Omega)}
R_{k-1}^{(c)}
\right\|_2^2,
\label{eq:app_practical_latent_score}
\end{equation}
where $c$ indexes the latent channels. This score measures whether the current
local latent region still contains directional residual variations. Regions with
larger $\widehat{D}_{k-1}(\Omega)$ are more likely to induce visible
fine-scale changes after decoding and are therefore retained for computation,
whereas regions with small scores are treated as spatially redundant.

In the proposed LHEP module, $\mathcal{H}$ is instantiated by the directional high-frequency components of the $2\times2$ Haar-DWT. This choice provides a minimal local finite-difference basis for estimating the high-frequency residual
energy in Eq.~\eqref{eq:app_practical_latent_score}, while avoiding explicit decoding at every scale.

\section{Haar-DWT High-frequency Energy and Kernel-size Analysis in LHEP}
\label{app:haar_proxy}

We first clarify how Haar-DWT computes the local high-frequency energy used in
LHEP. For a local $2\times2$ latent patch from one channel,
\begin{equation}
P =
\begin{bmatrix}
a & b \\
c & d
\end{bmatrix},
\end{equation}
the Haar basis decomposes $P$ into one low-frequency component and three
directional high-frequency components:
\begin{equation}
\begin{aligned}
z_{LL} &= \frac{1}{2}(a+b+c+d),\\
z_{LH} &= \frac{1}{2}(-a+b-c+d),\\
z_{HL} &= \frac{1}{2}(-a-b+c+d),\\
z_{HH} &= \frac{1}{2}(-a+b+c-d),
\end{aligned}
\end{equation}
where $z_{LL}$ denotes the local average component, and $z_{LH}$, $z_{HL}$, and $z_{HH}$ denote the horizontal, vertical, and diagonal high-frequency components, respectively. Since our LHEP aims to select regions that still require local refinement rather than reconstruct the full latent signal, we discard the low-frequency component and define the local high-frequency energy as
\begin{equation}
E_{\mathrm{HF}}(P)
=
z_{LH}^{2}+z_{HL}^{2}+z_{HH}^{2}.
\end{equation}
For a multi-channel latent feature map, the same decomposition is applied to each
channel and the energy is accumulated across channels:
\begin{equation}
\mathbf{E}_{raw}(i,j)
=
\sum_{c=1}^{C}
\sum_{d\in\{LH,HL,HH\}}
\left(
\mathbf{R}_{k-1}^{(c)} * \mathbf{K}_{d}
\right)_{i,j}^{2}.
\end{equation}
This corresponds to Eq.~\eqref{eq:raw_energy} in the main paper. The resulting
energy map measures whether each local latent patch still contains directional
residual variations that may lead to visible fine-scale refinement after
decoding.

Table~\ref{tab:haar_kernel_size} further analyzes the kernel size of the
Haar-based high-frequency operator used in LHEP. Under the same speedup of
$1.64\times$, the fixed $2\times2$ filters achieve the best overall quality
across GenEval, PSNR, SSIM, and LPIPS. Increasing the kernel size to
$4\times4$, $6\times6$, or $8\times8$ does not bring additional benefits,
indicating that larger kernels are less favorable for preserving localized
high-frequency responses. These results support our design choice that, for
LHEP-based region selection in late-stage VAR generation, a minimal local Haar
basis is sufficient to capture directional residual variations while preserving
well-localized refinement cues.

\begin{table}[t]
\centering
\caption{
Kernel-size ablation for Haar-based latent high-frequency energy on Infinity-2B.
}
\label{tab:haar_kernel_size}
\begin{tabular}{lccccc}
\toprule
Method & GenEval $\uparrow$ & PSNR $\uparrow$ & SSIM $\uparrow$ & LPIPS $\downarrow$ & Speedup $\uparrow$ \\
\midrule
$8\times8$ & 0.7311 & 23.9812 & 0.8749 & 0.1352 & 1.64$\times$ \\
$6\times6$ & 0.7309 & 26.1954 & 0.8751 & 0.1355 & 1.64$\times$ \\
$4\times4$ & 0.7320 & 26.1759 & 0.8746 & 0.1359 & 1.64$\times$ \\
$2\times2$ & \textbf{0.7334} & \textbf{27.3853} & \textbf{0.8769} & \textbf{0.1348} & 1.64$\times$ \\
\bottomrule
\end{tabular}
\end{table}

\begin{figure*}[t]
\centering
\includegraphics[width=1.\textwidth]{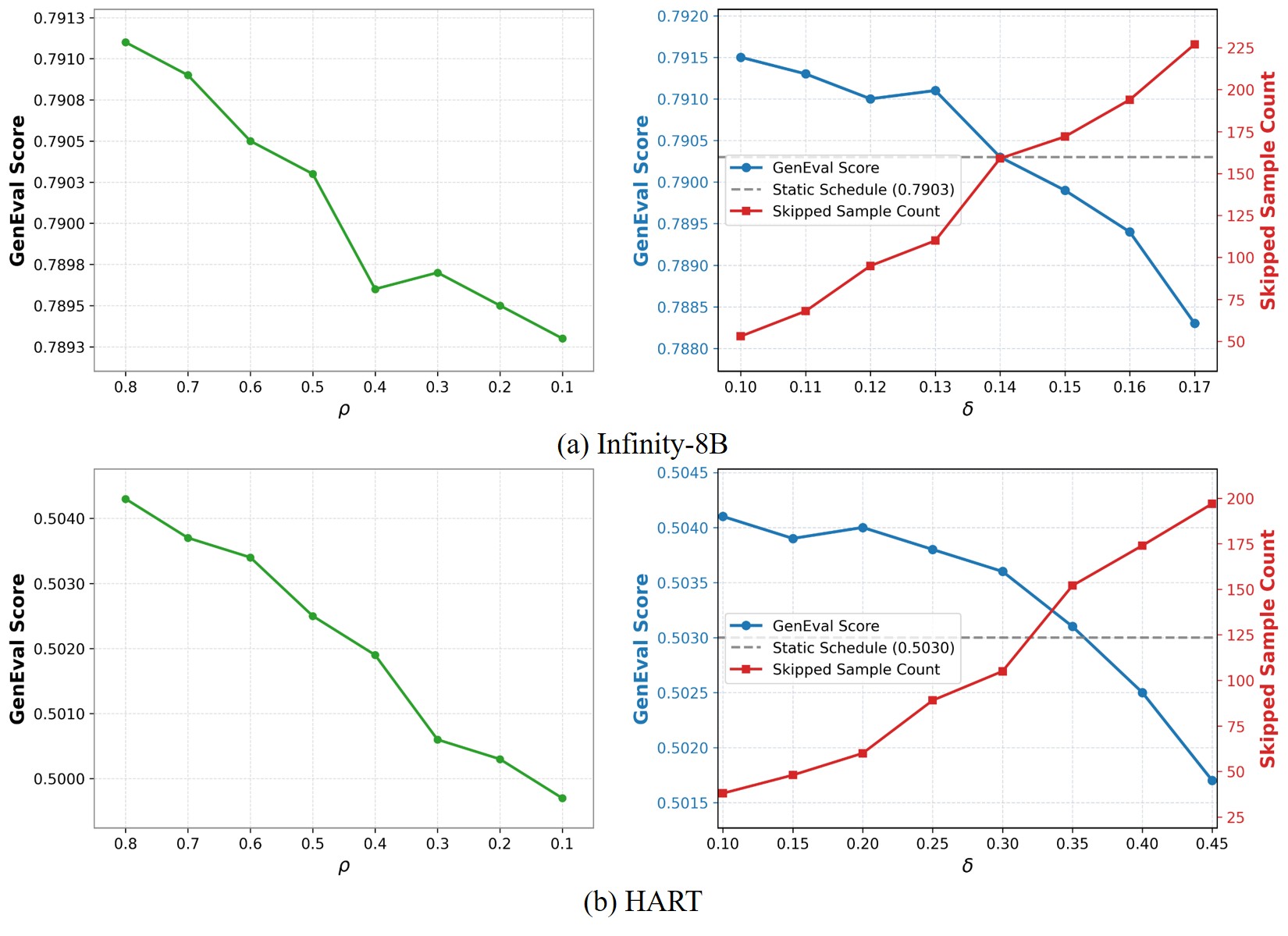}
\caption{
Hyperparameter sensitivity analysis of LD-Pruning on Infinity-8B and HART.
Left: sensitivity to the LHEP energy preservation ratio $\rho$.
Right: sensitivity to the SATS convergence tolerance $\delta$, where the blue
curve denotes GenEval score and the red curve denotes the number of skipped
samples.
}
\label{fig:hyper_sensitivity}
\end{figure*}

\section{Additional Hyperparameter Sensitivity Analysis}
\label{app:hyper_sensitivity}

We provide additional hyperparameter sensitivity analysis for the two key
hyperparameters in LD-Pruning: the energy preservation ratio $\rho$ in LHEP and
the convergence tolerance $\delta$ in SATS. The parameter $\rho$ controls the
sparsity of the spatial pruning mask, while $\delta$ controls how aggressively
SATS terminates the unconditional branch.

As shown in Figure~\ref{fig:hyper_sensitivity}, both Infinity-8B and HART exhibit
stable performance within a reasonably wide range of hyperparameters. For $\rho$,
the GenEval score changes smoothly as the pruning ratio varies. Performance
remains stable when $\rho$ is set within a moderate range, while overly aggressive
pruning leads to a gradual decrease in generation quality. This indicates that
LHEP is not highly sensitive to the exact choice of $\rho$ and only requires mild
tuning to balance quality and efficiency.

For $\delta$, increasing the threshold leads to more skipped samples, improving
efficiency but gradually reducing GenEval when the termination becomes too
aggressive. Nevertheless, both backbones maintain comparable performance to the
static schedule over a broad range of $\delta$. These results suggest that SATS
is robust to the threshold choice and can achieve a favorable speed--quality
trade-off without careful model-specific tuning. Overall, the sensitivity analysis
demonstrates that LD-Pruning is not strongly dependent on hyperparameter choices
and can be transferred across different VAR backbones with only minor adjustment.

\section{Additional Qualitative Results on Visual Fidelity}
\label{sec:appendix_visual}
\begin{figure}[t]
\begin{center}\centerline{\includegraphics[width=1.0\textwidth]{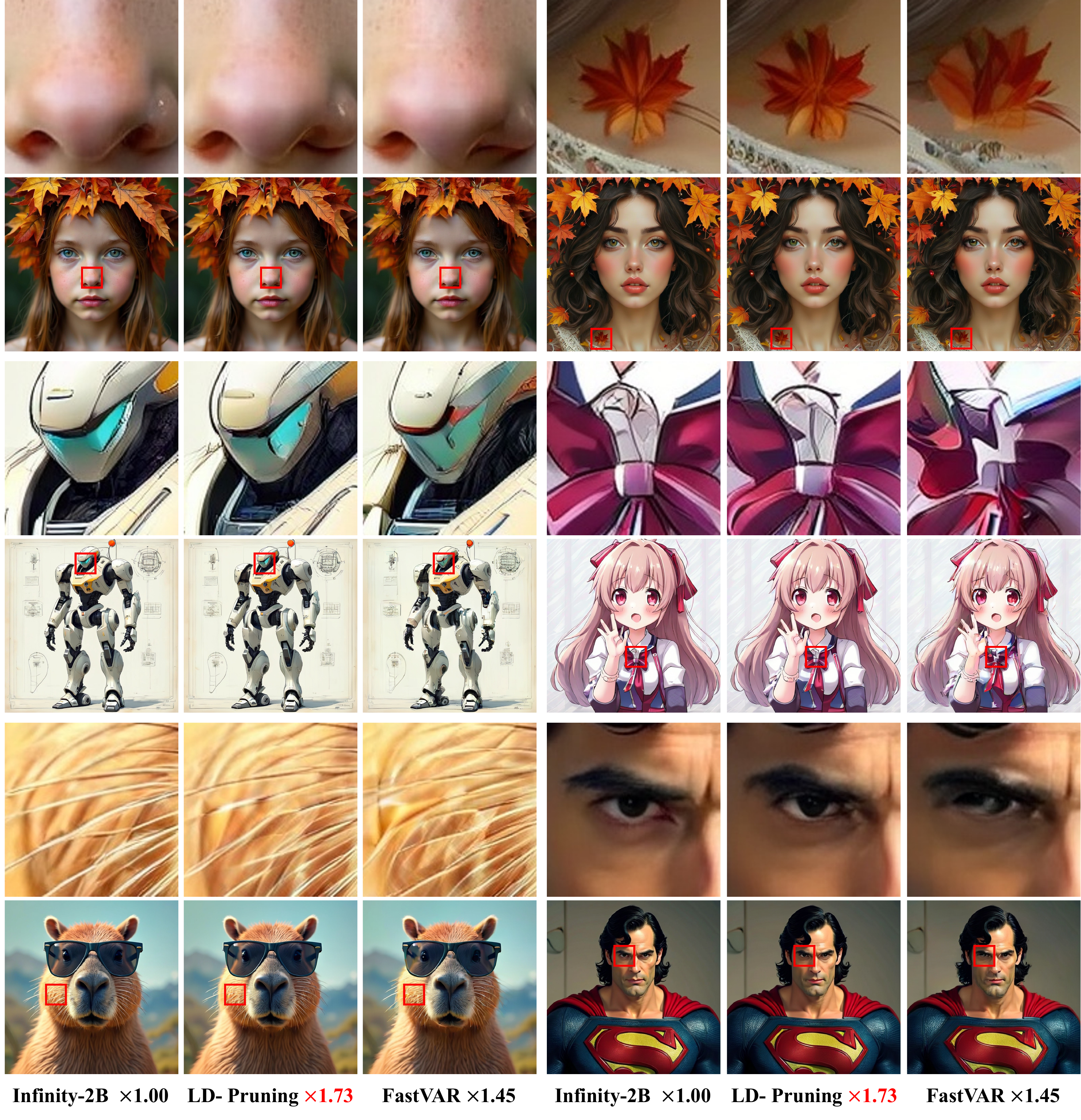}}
    \caption{Qualitative comparison of complex scene generation on HPSv2.1 \cite{wu2023human} benchmark. Zoom in for fine-detail visualization.
    }
    \label{visual-2B}
  \end{center}
  \vskip -0.4in
\end{figure}

\begin{figure}[t]
\begin{center}\centerline{\includegraphics[width=1.0\textwidth]{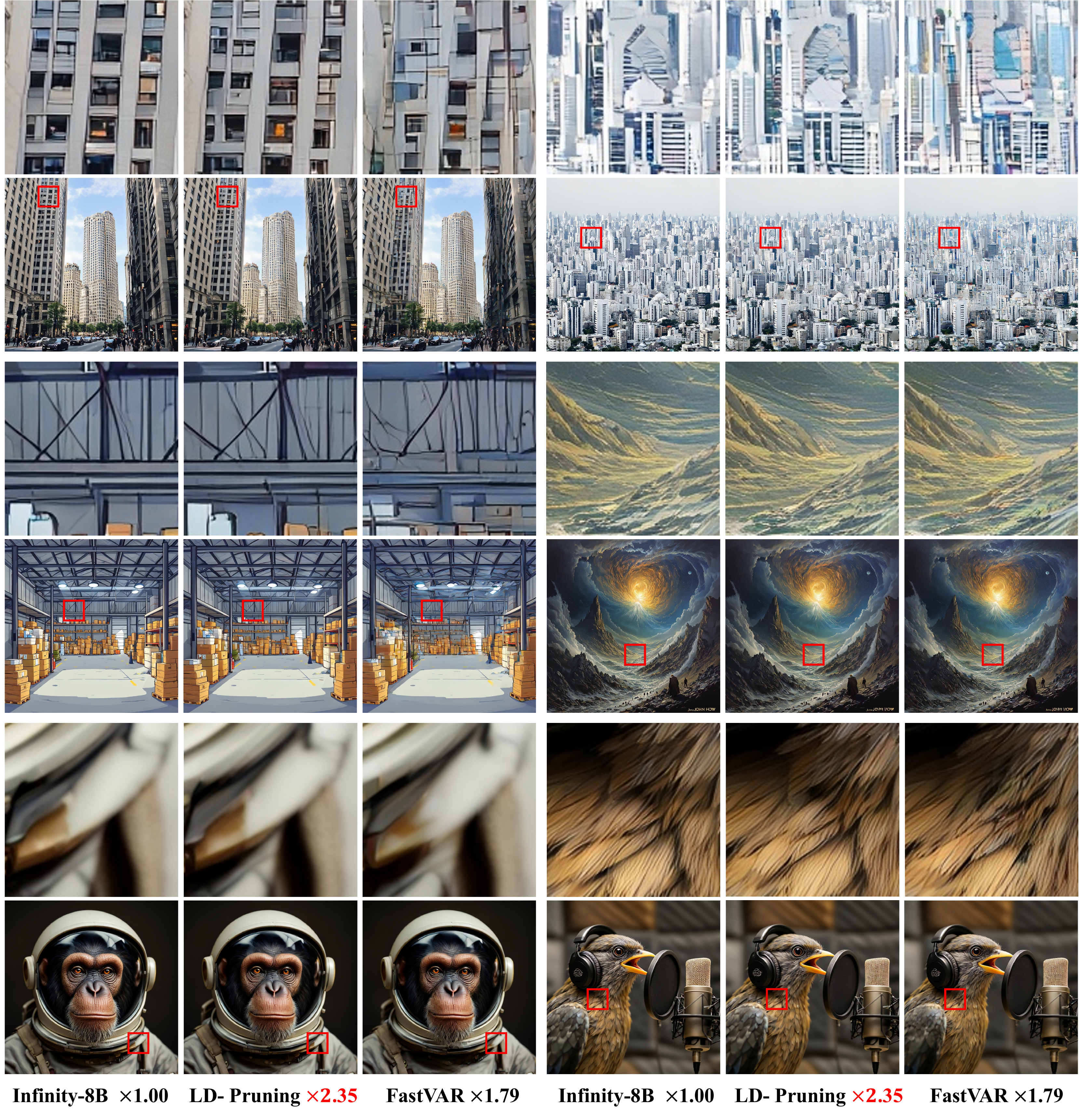}}
    \caption{Qualitative comparison of complex scene generation on HPSv2.1 \cite{wu2023human} benchmark. Zoom in for fine-detail visualization.
    }
    \label{visual-8B}
  \end{center}
  \vskip -0.4in
\end{figure}

Figure \ref{visual-2B} and Figure \ref{visual-8B} report qualitative comparisons on the HPSv2.1 benchmark, with zoom-in views for fine-detail inspection.

\paragraph{Fine-detail preservation.}
Across diverse complex scenes, LD-Pruning better preserves high-frequency textures in the zoomed regions, such as thin structures, subtle edges, and dense textured areas (e.g., hair/fur-like patterns or fine fabric/contour details). This observation is consistent with our main-body claim that LD-Pruning retains high-frequency textures in challenging regions where heuristic pruning baselines tend to blur or lose fine structures.

\paragraph{Comparison to fixed-ratio token pruning.}
In contrast, FastVAR—using a fixed pruning behavior—often yields over-smoothed details or weakened local contrasts in the same zoomed regions, especially when the scene contains intricate semantics and cluttered textures (Figure \ref{visual-2B}/\ref{visual-8B}). These qualitative results corroborate that identifying redundancy purely from earlier, semantically entangled representations can be unreliable for texture-heavy regions, whereas LD-Pruning—guided by latent discrepancy closer to the image/pixel domains—more reliably avoids pruning tokens that are crucial for visual fidelity.

\section{Algorithm Detail of LD-Pruning}
\label{app:pseudocode}

Algorithm~\ref{alg:ldpruning_2loop} summarizes the inference procedure of \textbf{LD-Pruning}. It takes a pretrained VAR model, the input prompt, and inference hyperparameters (e.g., target scales and guidance scale) as input, and outputs the final image.
At each generation scale, we compute the \emph{latent discrepancy} from intermediate representations as an online signal to quantify the marginal impact of computation on the pixel-domain output.
Guided by this signal, LD-Pruning performs two complementary actions:
(i) \textbf{LHEP} keeps a sparse set of tokens at detail generation scales and prunes the rest, while recovering pruned positions using cached/upsampled features;
(ii) \textbf{SATS} detects stable convergence (i.e., sufficiently small discrepancy) and early-terminates subsequent refinement to avoid redundant computation.
Overall, LD-Pruning enables adaptive, sample-wise acceleration without a fixed skipping schedule, and integrates into standard VAR inference by applying pruning/termination only at designated scales.

\begin{algorithm}[t]
\caption{LD-Pruning Inference for VAR with CFG}
\label{alg:ldpruning_2loop}
\textbf{Input:} scale steps $\{1,2,\ldots,K\}$ with resolutions $(h_k,w_k)$; semantic/outline scales $\{1,\ldots,K-m\}$; detail scales $\{K-m+1,\ldots,K\}$; \\
\hspace*{1.1em} VAR model $\phi$; quantizer $Q$; decoder $\mathcal{D}$; CFG logits $l^{(k)}_{\text{cond}}, l^{(k)}_{\text{uncond}}$ (when computed); \\
\hspace*{1.1em} SATS scales $\mathcal{S}_{\text{SATS}}$ (e.g., $\{9,10,11,12\}$); LHEP scales $\mathcal{S}_{\text{LHEP}}$ (e.g., $\{10,11,12\}$); thresholds $\delta,\rho$; $\epsilon$ for stability.\\
\textbf{Output:} final generated image $I$.

\begin{algorithmic}[1]
\STATE $F_0 \leftarrow 0$; \quad $R_0 \leftarrow \langle SOS\rangle$; \quad $\texttt{stopUB}\leftarrow\textbf{false}$; \quad $\Delta_0\leftarrow 0$

\vspace{0.2em}
\STATE \textbf{// (I) Semantic \& structure establishment stages (standard VAR)}
\FOR{$k = 1,2,\ldots,K-m$}
    \STATE $\tilde{F}_{k-1} \leftarrow Down(F_{k-1},(h_{k-1},w_{k-1}))$
    \STATE $F^{o}_{k} \leftarrow \phi(\tilde{F}_{k-1})$
    \STATE $R_k \leftarrow Q(F^{o}_{k})$
    \STATE $F_k \leftarrow F_{k-1} + Up(R_k,(h_K,w_K))$
\ENDFOR

\vspace{0.2em}
\STATE \textbf{// (II) Detail refinement stages: SATS + LHEP + sparse update}
\FOR{$k = K-m+1,\ldots,K$}

    \IF{$(k \in \mathcal{S}_{\text{SATS}})\ \land\ (\texttt{stopUB}=\textbf{false})$}
        \STATE $\Delta_k \leftarrow \|l^{(k)}_{\text{cond}} - l^{(k)}_{\text{uncond}}\|_2$
        \STATE $\gamma_k \leftarrow \frac{|\Delta_k-\Delta_{k-1}|}{\Delta_{k-1}+\epsilon}$
        \IF{$\gamma_k < \delta$}
            \STATE $\texttt{stopUB} \leftarrow \textbf{true}$ \COMMENT{stop unconditional branch for subsequent detail scales}
        \ENDIF
    \ENDIF

    \IF{$k \in \mathcal{S}_{\text{LHEP}}$}
        \STATE $M \leftarrow \textsc{LHEP\_Mask}(R_{k-1}, \rho)$ \COMMENT{latent HF energy $\rightarrow$ dynamic mask}
    \ELSE
        \STATE $M \leftarrow \mathbf{1}$
    \ENDIF

    \STATE $\tilde{F}_{k-1} \leftarrow Down(F_{k-1},(h_{k-1},w_{k-1}))$
    \STATE $\tilde{F}^{r}_{k} \leftarrow \textsc{Gather}(\tilde{F}_{k-1}, M)$ \COMMENT{active tokens only}
    \STATE $F^{r}_{k} \leftarrow \phi(\tilde{F}^{r}_{k})$ \COMMENT{sparse inference; UB used only if $\neg$\texttt{stopUB}}
    \STATE $R^{r}_{k} \leftarrow Q(F^{r}_{k})$

    \STATE $R_{\text{base}} \leftarrow Up(R_{k-1},(h_k,w_k))$ \COMMENT{reuse previous residual on skipped regions}
    \STATE $R_k \leftarrow \textsc{Scatter}(R_{\text{base}}, R^{r}_{k}, M)$
    \STATE $F_k \leftarrow F_{k-1} + Up(R_k,(h_K,w_K))$
\ENDFOR

\STATE $I \leftarrow \mathcal{D}(F_K)$
\STATE \textbf{return} $I$

\vspace{0.4em}
\STATE \textbf{Subroutine:} \textsc{LHEP\_Mask}$(R_{k-1},\rho)$
\STATE $E_{\text{raw}} \leftarrow \sum_{d\in\{LH,HL,HH\}}\|R_{k-1} * K_d\|_2^2$ \COMMENT{Haar-DWT energy}
\STATE $E_{\text{target}} \leftarrow \textsc{Interpolate}(\textsc{AvgPool}(E_{\text{raw}}))$
\STATE $v \leftarrow \textsc{sort}(\textsc{vec}(E_{\text{target}}))$ (desc); find smallest $k^\star$ s.t. $\sum_{i=1}^{k^\star} v_i \ge \rho\|v\|_1$
\STATE $\tau \leftarrow v_{k^\star}$; \quad $M(i,j)\leftarrow \mathbb{I}[E_{\text{target}}(i,j)\ge \tau]$
\STATE \textbf{return} $M$
\end{algorithmic}
\end{algorithm}

%%%%%%%%%%%%%%%%%%%%%%%%%%%%%%%%%%%%%%%%%%%%%%%%%%%%%%%%%%%%%%%%%%%%%%%%%%%%%%%
%%%%%%%%%%%%%%%%%%%%%%%%%%%%%%%%%%%%%%%%%%%%%%%%%%%%%%%%%%%%%%%%%%%%%%%%%%%%%%%

\end{document}